\documentclass[sigconf]{acmart}
\usepackage[utf8]{inputenc}
\usepackage[T1]{fontenc}

\AtBeginDocument{%
  \providecommand\BibTeX{{%
    \normalfont B\kern-0.5em{\scshape i\kern-0.25em b}\kern-0.8em\TeX}}}

\setcopyright{acmcopyright}
\copyrightyear{2020}
\acmYear{2020}
\acmDOI{}

\begin{document}

\title{Capturing Evolution in Word Usage: Just Add More Clusters?}



\author{Matej Martinc}
\authornote{The authors contributed equally to this research.}
\email{matej.martinc@ijs.si}
\affiliation{%
  \institution{Jozef Stefan Institute}
  \country{Slovenia}
}

\author{Syrielle Montariol}
\authornotemark[1]
\email{syrielle.montariol@limsi.fr}
\affiliation{\institution{LIMSI - CNRS, Univ. Paris-Sud, Univ. Paris-Saclay \\
Societ\'e G\'en\'erale
}
\country{France}}

\author{Elaine Zosa}
\authornotemark[1]
\email{elaine.zosa@helsinki.fi}
\affiliation{%
  \institution{University of Helsinki}
  \country{Finland}}

\author{Lidia Pivovarova}
\email{lidia.pivovarova@helsinki.fi}
\affiliation{%
  \institution{University of Helsinki}
  \country{Finland}}

\renewcommand{\shortauthors}{Martinc, et al.}

\begin{abstract}
 The way the words are used evolves through time, mirroring cultural or technological evolution of society. Semantic change detection is the task of detecting and analysing word evolution in textual data, even in short periods of time. In this paper we focus on a new set of methods relying on contextualised embeddings, a type of semantic modelling that revolutionised the NLP field recently. We leverage the ability of the transformer-based BERT model to generate contextualised embeddings capable of detecting semantic change of words across time. Several approaches are compared in a common setting in order to establish strengths and weaknesses for each of them. We also propose several ideas for improvements, managing to drastically improve the performance of existing approaches. 
 
\end{abstract}

\ccsdesc[500]{Computing methodologies~Lexical semantics}
\ccsdesc[300]{Computing methodologies~Cluster analysis}
\ccsdesc[300]{Information systems~Language models}

\keywords{Semantic Change, Contextualised Embeddings, Clustering}

\maketitle

\section{Introduction}

The large majority of data on the Web is unstructured. Amongst it, textual data is an invaluable asset for data analysts. With the large increase in volume of interaction and overall usage of the Web, more and more content is digitised and made available online, leading to a huge amount of textual data from many time periods becoming accessible. However, textual data are not necessarily homogeneous as they rely on a crucial element that evolves throughout time: language. Indeed, a language can be considered as a dynamic system where word usages evolve over time, mirroring cultural or technological evolution of society~\cite{Aitchison}.

For example, the word {\em domain} has seen its usage shifting these past years, from the sense of "an area or territory owned by a government", towards the sense of "a subset of the Internet."

In linguistics, \textit{diachrony} refers to the study of such temporal variations in the use and meaning of a word. While analysing textual data from the Web, detecting and understanding these changes can be done for two types of usages. On the one hand, it can be used directly for linguistic research or social analysis, by interpreting the reason of the semantic change and linking it to real-world events, and by analysing trends, topics and opinions evolution \cite{GillaniDynamicPerception19}. On the other hand, it can be used as a support for many tasks in Natural Language Processing (NLP), from text classification to information retrieval conducted on a temporal corpora where semantic change might occur.

To tackle semantic change, models usually rely on word embeddings, such as Word2Vec~\cite{mikolov2013efficient} and Glove~\cite{Pennington14GLOVE}, which summarise all senses and usages of a word into one vector at one point in time. Measuring the distance between these vectors across time periods is used to detect and quantify the differences in meaning. But these methods do not take into consideration that most words have multiple senses, since all word usages are aggregated into a single static word embedding. Contextualised embeddings models such as BERT~\cite{devlin2018bert} are capable of generating a different vector representation for each specific word usage, making them more suitable for this task.

The goal of this paper is to establish the best way to detect semantic change in a temporal corpus by capitalising on BERT contextualised embeddings. First, several approaches for semantic shift detection from the literature are compared in a common setting in order to establish strengths and weaknesses of each specific method. Second, several improvements are presented, which manage to drastically improve the performance of existing approaches. Our code and models will be made publicly available.

\section{Related Work}

A large majority of methods for semantic shift detection leverage dense word representations (embeddings). Word-frequency methods for detecting semantic shift that were used in earlier studies~\cite{juola2003time,hilpert2008assessing}, are now rarely used.  The detailed overview of the field could be found in recent surveys~\cite{kutuzov2018diachronic, tahmasebi2018survey, tang_2018}.

\subsection{Static Word Embeddings for Semantic Change}

The first research that employed word embeddings for semantic shift detection was conducted by~\cite{kim2014temporal}, who leveraged a continuous Skipgram model proposed by \cite{mikolov2013efficient}. The main idea was to train a separate embedding model for each time period. Since embedding algorithms are inherently stochastic and the resulting embedding sets are invariant under rotation, a procedure that makes these models comparable is needed. To solve this problem, they proposed the incremental model fine-tuning approach, where the weights of the model, trained on a certain time period, are used to initialize weights of a model trained on the next successive time period. Some improvements of the approach were later proposed by~\cite{peng2017incrementally}, who replaced the softmax function for the continuous skipgram model with a more efficient hierarchical softmax, and by ~\cite{kaji2017incremental}, who proposed an incremental extension for negative sampling.

An alternative approach was proposed in~\cite{kulkarni2015statistically}, where embedding models trained on different time periods were aligned in a common vector space after the initial training using a linear transformation for the alignment. The approach was upgraded~\cite{zhang2016past} by using a set of nearest neighbour words as anchors for the alignment. 

A third alternative for semantic shift detection with static word embeddings is to treat the same words in different time periods as different tokens in order to get time specific word representations for each time period~\cite{rosenfeld2018deep,dubossarsky2019time}. Here, only one embedding model needs to be trained and no aligning is needed.

\subsection{The Emergence of Contextualised Embeddings}

While in the traditional static word embedding models each word from the predefined vocabulary is presented as a unique vector, here a different vector is generated for each context a word appears in, i.e., for each word usage.  The two most widely used contextual embeddings models are ELMo (Embeddings from LanguageModels~\cite{peters2018deep}) and a more recent BERT (Bidirectional Encoder Representations from Transformers~\cite{devlin2018bert}). 

The approach of using contextual embeddings for semantic shift detection is fairly novel; we are aware of three recent studies that employed it.

It was first used in a controlled way \cite{hu-etal-2019-diachronic}: for a set of polysemic words, a representation for each sense is learned using BERT. Then pretrained BERT is applied to a diachronic corpus, extracting token embeddings, that are matched to the closest sense embedding. Finally, the proportions for each sense are computed at each successive time slice, revealing the evolution of the distribution of senses for each target word. This method requires that the set of senses of each target word is known beforehand. 

Another possibility is clustering all contextual embeddings for a target word into clusters representing the word senses or usages in a specific time periods~\cite{giulianellilexical}. K-means clustering and BERT contextual embeddings were used in this study.  In addition, the incremental training approach proposed by~\cite{kim2014temporal} was used for diachronic fine-tuning of the model. To quantify changes between word usages in different time periods, Jensen-Shannon divergence(JSD), a measure of similarity between probability distributions, was used. They also tested if domain adaptation of the model would improve the results of their approach by fine-tuning the model on an entire corpus rather than on specific time periods, however this yielded no performance improvements.

In the third, even more recent study, contextual embeddings for a specific word in a specific time period were averaged in order to generate a time specific word representation for each word in each period~\cite{martinc2019leveraging}. BERT embeddings are used in the study and cosine distance is used for measuring the difference between word representations in different time periods. 

\section{Data}
\label{sec:data}

For the experiments, we use the Corpus of Historical American English (COHA) \footnote{https://www.english-corpora.org/coha/}. It contains more than 400 million words of text from the 1810s-2000s. As a historical corpus, it is smaller than the widely used Google books corpus \footnote{http://googlebooks.byu.edu/} but it has the advantage of being balanced by genre (fiction, magazines, newspapers, and non-fiction texts, gathered from various Web sources) for each decade. We focus our experiments on the most recent data in this corpus, from the 1960s to the 1990s (1960s has around 2.8 million and 1990s 3.3 million words), to match the evaluation corpus. The fine-tuning of the model is also done only on this subset. 

We rely on a small human-annotated dataset~\cite{Gulordava} to conduct the evaluation.  The dataset consists of 100 words from various frequency ranges, labelled by five annotators according to the level of semantic change between the 1960s and the 1990s. They use a 4-points scale from "0: no change" to "3: significant change", and the inter-rater agreement was 0.51 (p <0.01, average of pair-wise Pearson correlations).  The most significantly changed words from the dataset are, for example, {\em user} and {\em domain}; words for which the meaning remain intact, are for example {\em justice} and {\em chemistry}.  
This dataset is a valuable resource and has been used to evaluate methods for measuring semantic change in previous research~\cite{frermann2016bayesian,giulianellilexical}.

Following previous work, we use the average of the human annotations as semantic change score. For evaluation, we compute Pearson and Spearman rank correlations between a model output and this score.  The notion of the best model is based on Spearman correlations.

\section{Methodology}\label{sec:model}

\subsection{Context-dependent Embeddings}

BERT is a neural model based on the transformer architecture first proposed by \cite{vaswani2017attention}. It relies on a transfer learning approach proposed by \cite{howard2018universal}, where in the first step the network is pretrained as a language model on large corpora in order to learn general contextual word representations. This is usually followed by a task specific fine-tuning step e.g., classification or, in our case, domain adaptation. BERT's novelty is an introduction of a new pretraining learning objective, a \textit{masked language model}, where a percentage of words from the input sequence is masked in advance, and the objective is to predict these masked words from an unmasked context. This allows BERT to leverage both left and right context, meaning that a word $w\textsubscript{t}$ in a sequence is not determined just from its left sequence $w\textsubscript{1:{t-1}} = [w_1,...,w_{t-1}]$---as is the case in the traditional language modelling task---but also from its right word sequence $w\textsubscript{t+1:n} = [w_{t+1},...,w_{t+n}]$. 


In our experiments we use the English BERT-base-uncased model with 12 attention layers and a hidden layer of size 768, which was pretrained on the Google Books Corpus \citep{goldberg2013dataset} (800M  words) and Wikipedia (2,500M words). For some of the experiments (see Table \ref{tab:results}), we further fine-tune this model (as a \textit{masked language model}) for up to 10 epochs on the COHA subcorpus described in Section \ref{sec:data} for domain adaptation. 

Note that our fine-tuning approach deviates from the approaches presented in some of the related work \cite{giulianellilexical} and we do not conduct any diachronic fine-tuning of the model using the incremental training approach similar to~\cite{kim2014temporal}. The hypothesis is that this step is not necessary due to contextual nature of embeddings generated by the model, which  by definition are dependent on the context that is always time-specific.

In order to acquire a contextual embedding for each token in the corpus, the temporal corpus documents are first split into 256 tokens long sequences of byte-pair encodings \cite{wu2016google} and fed into a BERT model. A sequence embedding is generated for each of these sequences by summing last four encoder output layers. Finally, this sequence embedding of size \textit{sequence length x embeddings size} is cut into pieces, to get a separate contextual embedding for each token in the sequence.

\subsection{Target Words Selection} \label{sec:prelim_measures}

The evaluation dataset that we use in this paper includes 100 pre-selected words, so we are able to apply clustering for each of them. However in any practical application of semantic change detection, performing clustering for every word in the corpus would not be feasible in terms of computing time. Thus, we investigate a preliminary step to select a set of words that may have undergone semantic change.

We use several scalable metrics, that we compare in Section~\ref{sec:prelim-exp}. A first set of metrics relies on the computation of a variation measure: for each word, similarly to \cite{kutuzov2020diachronic}, we compute the average token embedding on the full corpus.   Then, we compute the cosine distance between each token embedding and this average embedding. Finally, we take the mean of all these cosine distances as the \textit{variation coefficient} of the word. 

We also use a measure of evolution of a word's variation. Dividing our corpus into decades, we compute the variation coefficient inside each time slice $Var_t$. Then, we take the average difference from one time step to another:
\begin{equation*}
    \text{Variation by time slice} = \text{mean} (\text{Variation}_t - \text{Variation}_{t-1})
\end{equation*}

The second set of metrics relies on \textit{averaging} the token embeddings by time slice, and using the cosine distance as a measure of semantic drift between time slices. The total drift is the cosine distance between the average of token representations of the first time slice and of the last time slice. It represents the amount of change a word has undergone from the first to the last period, without using the other time slices, so without taking into account the variations in between. The \textit{averaging by time slice} computes the mean of the drifts from each time step to the next one.

We select a threshold (as a fraction of the size of the full vocabulary) to get a list of target words. Then, the heavier clustering techniques can be applied to this list.

\subsection{Embeddings Clustering}


The goal of the clustering step is to group the word occurrences by similar vector representation. Then we can compare cluster distribution across time periods to detect semantic shift by using JSD, same as in \cite{giulianellilexical}.
The intuition is the following: if, for instance, a word acquired a novel sense in the latter time period, then a cluster corresponding to this sense only consists of word usages from this period but not the earlier ones, which would be reflected by a higher divergence.

However, 
one cluster does not necessarily correspond to a precise sense of the word. Each cluster would rather represent a specific usage or context.  Moreover, a word may completely change its context without changing the meaning. Consequently, determining the number of clusters is a tricky part.

For clustering we used k-means with various values for $k$ and affinity propagation~\cite{frey2007clustering}.  Affinity propagation has been previously used for various linguistic tasks, such as word sense induction~\cite{alagic2018leveraging,kutuzov2017clustering}.  Affinity propagation is based on incremental graph-based algorithm, partially similar to PageRank. Its main strength is that number of clusters is not defined in advance but inferred during training.  In this work we used a Scikit-learn implementation with standard hyperparameters.

We also experiment with the approach inspired by~\cite{Amrami-WSI}, where clusters with less than two members are considered weak clusters and are merged with the closest strong cluster, i.e. clusters with more than two members.\footnote{Note that procedure in~\cite{Amrami-WSI} is more complex: they first find one or more number of representatives for each datapoint and then clustering is applied over representatives, while in our work clustering is done over the instances themselves.} We refer to this method as two-stage clustering. For the initial clustering step, we also used k-means or affinity propagation.



\section{Experiments}

We focus our analysis on comparing the various clustering approaches and the metrics to detect semantic change. For our experiments, we use a pretrained version of BERT \footnote{https://pytorch.org/hub/huggingface\_pytorch-transformers/}. BERT fine-tuned on the COHA subcorpus for up to 10 epochs is also tested. For clustering, we make use of the Scikit-learn implementation of k-means and affinity propagation \footnote{https://scikit-learn.org/stable/modules/clustering.html}. For k-means, we set the number of clusters $k$ and use default parameters for the rest. Similarly, for affinity propagation, we use the default parameters set by the library.

For evaluation, we rely on the 100 gold standard annotated drifts presented in section \ref{sec:data}. However, a specificity of BERT is the representation of words with byte-pair encodings \cite{wu2016google}. Thus, some words can be divided into several sub-parts; for example, in our list of target words for evaluation, \textit{sulphate} is divided into two byte-pairs \textit{sul} and \textit{\#\#phate}, where \textit{\#\#} denotes the splitting of the word. This is also true for the words \textit{mediaeval}, \textit{extracellular} and \textit{assay}. We decided to exclude these words from our analysis. 
Thus, our results are not directly comparable with the other approaches in the literature. 


\subsection{Preliminary Step: Target Word Selection} \label{sec:prelim-exp}

Table \ref{tab:results_prelim} shows the Pearson and Spearman correlations between the various preliminary drift measures defined in section \ref{sec:prelim_measures}, and the human-annotated drifts. All metrics are computed using the full dataset from the 1960s to the 1990s, except \textit{averaging} which is only computed between the first decade (1960s) and the last one (1990s) without using the textual data in between.

The measure with the highest correlation with the human annotations is averaging, which measures semantic drift between the first and the last time step. This intuitively makes sense as it resembles the way the evaluation dataset was annotated by only considering the first and the last decade. Variation by decade also shows good results; it is a measure of the evolution of the level of variation of a word usage through time.

\begin{table}[!t]
\begin{tabular}{|l|c|c|}
\hline
\multicolumn{1}{|c|}{\textbf{Method}} & \textbf{Pearson} & \textbf{Spearman} \\ \hline
Variation               & 0.070                        & 0.015                         \\ \hline
Variation by decade     & 0.239                        & 0.303                         \\ \hline
\textbf{Averaging}                  & \textbf{0.354}               & \textbf{0.349}                \\ \hline
Averaging by decade                & 0.295                        & 0.272                         \\ \hline
\end{tabular}
\caption{Correlations between drift measures and manually annotated list of semantic drifts~\cite{Gulordava} between 1960s and 1990s.}
\label{tab:results_prelim}
\vspace{-5mm}
\end{table}

\subsection{Clustering}
\label{sec:clusteri}

In Table~\ref{tab:results} we present the Pearson and Spearman correlation coefficients obtained using models described above. 

At the top of the table we overview previous work on the same testset. To train the models, \cite{Gulordava} used Google Books Ngrams, \cite{frermann2016bayesian} used an extended COHA corpus, and both~\cite{giulianellilexical} and~\cite{kutuzov2020diachronic} used a subcorpus of COHA, identical to the one used in our experiments.  In fact, the setting in~\cite{giulianellilexical} is quite similar to our work, though our best model performance is much higher than in~\cite{giulianellilexical}; we will further discuss this discrepancy in performance in Section \ref{sec:discussion}. 

As can be seen in Table~\ref{tab:results} the best result in our experiments is obtained using affinity propagation on the fine-tuned BERT model.  Results, obtained using pretrained and fine-tuned models are consistent: in both runs averaging yields lower performance than clustering and affinity propagation is the best clustering method. Two-stage clustering works better than k-means but slightly worse than affinity propagation.

Fine-tuning BERT improves all models except for k-means with 3 clusters and averaging---we do not yet have a clear explanation for that exception.

\begin{table}[!t]
\begin{tabular}{|l|r|r|}
\hline
\multicolumn{1}{|c|}{\textbf{Method}} & \textbf{Pearson} & 
\textbf{Spearman} \\ 
\hline
\multicolumn{3}{c}{\textbf{Related work}} \\ 
\hline
\textbf{Gulardova \& Baroni, 2011}~\cite{Gulordava}          & 0.386                        & -                           \\ 
\textbf{Frermann \& Lapata, 2016}~\cite{frermann2016bayesian}           & -                          & 0.377                         \\ 
\textbf{Giulianelli, 2019}~\cite{giulianellilexical}                  & 0.231                        & 0.293                         \\ 
\textbf{Kutuzov, 2020}~\cite{kutuzov2020diachronic}                 & 0.233                        & 0.285                         \\ 
\hline
\multicolumn{3}{c}{\textbf{Pretrained BERT}} \\ 
\hline
\textbf{Averaging}               & 0.354                      & 0.349                        \\ 
\hline
\textbf{k-means, k = 3}               & 0.461                        & 0.444                         \\ 
\textbf{k-means, k = 5}               & 0.476                        & 0.443                         \\ 
\textbf{k-means, k = 7}               & 0.485                        & 0.434                         \\ 
\textbf{k-means, k = 10}              & 0.478	                     & 0.443                       \\
\hline
\textbf{2-stage clustering, Aff. propagation}                       & 0.530                        & 0.485\\
\textbf{Affinity propagation}         & \textbf{0.548}               & \textbf{0.486}                \\ 

\hline
\multicolumn{3}{c}{\textbf{Fine-tuned BERT for 5 epochs}} \\ 
\hline
\textbf{Averaging}               & 0.317                      & 0.341                        \\ 
\hline
\textbf{k-means, k=3}   & 0.411 & 0.392 \\
\textbf{k-means, k=5}   & 0.539 & 0.508 \\
\textbf{k-means, k=7}   & 0.526 & 0.491 \\
\textbf{k-means, k=10}  & 0.500 & 0.466 \\
\hline
\textbf{2-stage clustering, Aff. propagation}                       & 0.554                        & 0.502                       \\ 
\textbf{Affinity propagation} & \textbf{0.560} & \textbf{0.510} \\
\hline
\end{tabular}
\caption{Correlations between detected semantic change and manually annotated list of semantic drifts~\cite{Gulordava} between 1960s and 1990s.}
\label{tab:results}
\vspace{-4mm}
\end{table}

\section{Discussion}
\label{sec:discussion}


\subsection{Error Analysis}

We manually checked few examples; we chose the words that have less mentions in the corpus to be able to look through all sentences containing the word.  One of the tricky cases for our model is the word {\em neutron}: according to the manual annotation, it is ranked 81st and has a stable meaning, while our best model considered it one of the most changed and ranked it at 9.  

\begin{figure}[t]
    \includegraphics[trim=0mm 18mm 0mm 20mm, clip,width=0.45\textwidth]{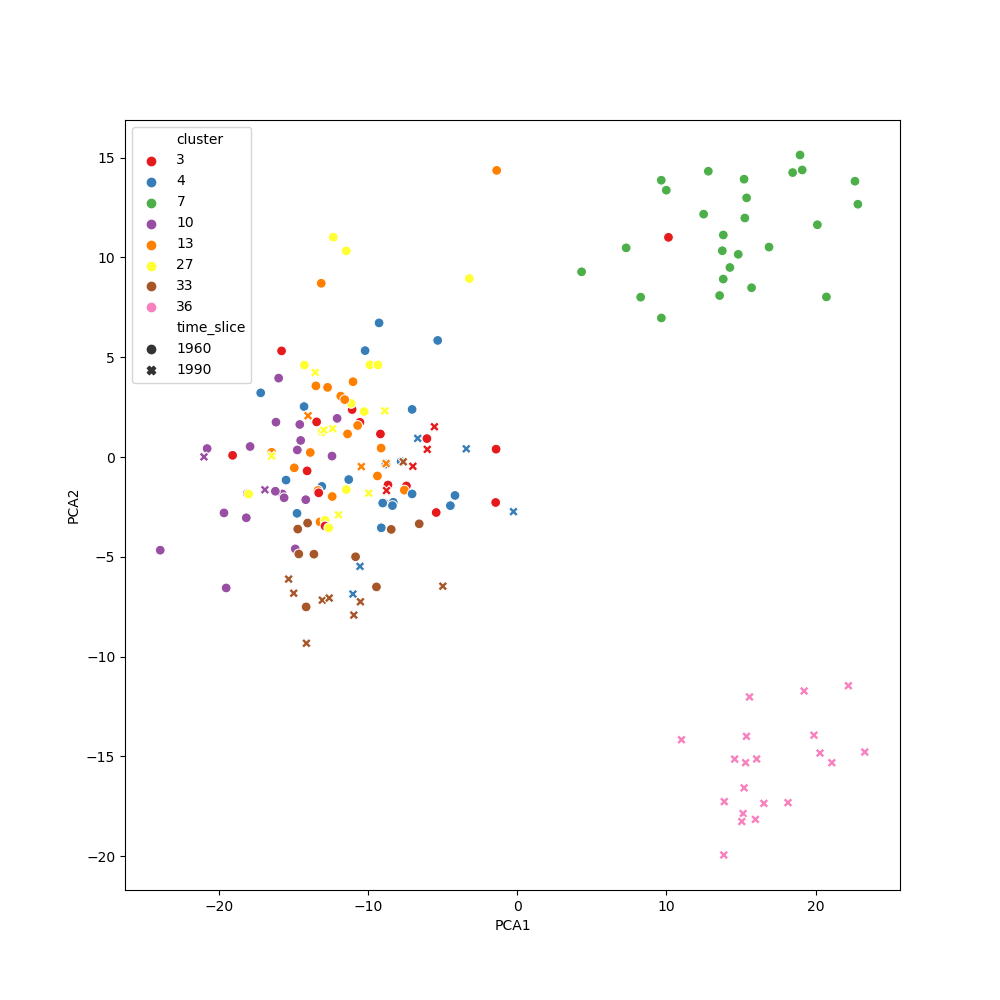}
    \caption{2D PCA visualization for the biggest clusters obtained for word {\em neutron}.}
    \label{fig:neutron}
    \vspace{-3mm}
\end{figure}

We visualize the biggest clusters for {\em neutron} using PCA decomposition of BERT embeddings (Figure~\ref{fig:neutron}). There are two clearly distinctive clusters: cluster 36 in the bottom right corner, drawn with pink crosses, which consists only of instances from 1990s, and cluster 7 drawn with green dots in the top right corner, which consists only of instances from 1960s.  A manual check reveals that the former cluster consists of sentences which mention {\em neutron stars}.  Though neutron stars have been already discovered in 1960s they were probably less known and are not represented in the corpus \footnote{\url{https://en.wikipedia.org/wiki/Neutron_star}}.  Nevertheless, a difference in a collocation frequency does not mean a semantic shift, since collocations often have a non-compositional meaning.  Thus, our method could be improved by removing stable multiword expressions from the training set.

The latter cluster, consisting of word usages from 1960s, contains many sentences that have a certain pathetic style and elevated emotions, such as underlined in the examples below: \\
{\em throughout the last several decades the \textunderscore{dramatic revelation} of this new world of matter has been dominated by a \textunderscore{most remarkable} subatomic particle  --  the neutron .} \\
{\em the discovery of the neutron by sir james chadwick in 1939. marked \textunderscore{a great step forward} in understanding the basic nature of matter .}

The lack of such examples in 1990s might have a socio-cultural explanation or it could be a mere corpus artefact.  In any case, this has nothing to do with semantic shift and demonstrates an ability of BERT to capture other aspects of language, including syntax and pragmatics.  

\subsection{Impact of Fine-tuning}

\begin{figure}[t]
    \includegraphics[trim=0mm 5mm 0mm 8mm, clip,width=0.4\textwidth]{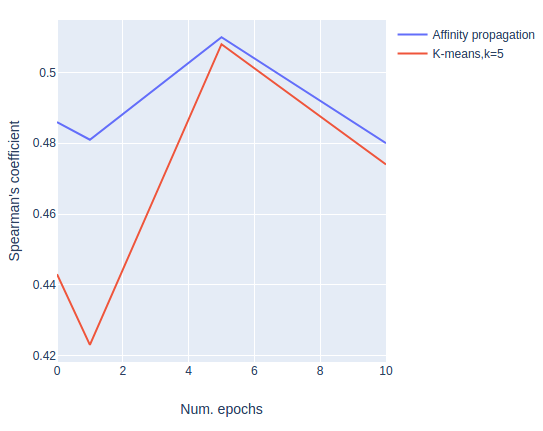}
    \caption{Impact of BERT fine-tuning on the performance of two distinct aggregation methods, affinity propagation and k-means with k=5.}
    \label{fig:fine_tuning}
    \vspace{-4mm}
\end{figure}

Figure \ref{fig:fine_tuning} shows the comparison of fine-tuning influence for two best clustering methods (affinity propagation, and k-means with k=5). Interestingly, a light fine-tuning (just for one epoch) decreases the performance of both methods (in terms of Spearman correlation) in comparison to no fine-tuning at all (zero epochs). After that, the length of fine-tuning until up to 5 epochs is linearly correlated with the performance increase.  


Fine-tuning the model for five epochs appears optimal. After that, the performance for both methods starts decreasing, most likely because of over-fitting due to the reduced size of the fine-tuning dataset compared to the training data.

The impact of fine-tuning on the k-means clustering is stronger than on the affinity propagation.  The difference between model's performance on 5 epochs is negligible.  However, this effect holds only with k=5, other values of k do not demonstrate such a difference between original and fine-tuned models, as can be seen in Table~\ref{tab:results}.

\subsection{Clustering}

When it comes to comparison between the semantic change detection approaches used in this work and the approaches used in the related works, most of the approaches proposed in this work manage to outperform previous approaches by a large margin. 

The proposed clustering approaches are methodologically very similar to the approaches proposed by \cite{giulianellilexical}, yet we manage to outperform their approach by a margin of about 35 percentage points when affinity propagation is used and by about 33 percentage points when k-means clustering\footnote{Here we are referring to our best k-means configuration with five clusters and using a BERT model fine-tuned for five epochs.}, same as in \cite{giulianellilexical}, is used. 

Unfortunately,~\cite{giulianellilexical} does not report a number of clusters that has been used, they only mention that the number of clusters has been optimised using the silhouette scores so we can only speculate why their results are much lower than ours. The first hypothesis is connected with the usage of the silhouette score, which might not be optimal for our goals.  We compute the silhouette score\footnote{Using standard Scikit-learn implementation, \url{https://scikit-learn.org/stable/modules/clustering.html\#silhouette-coefficient}} for clusterings obtained by our methods.  In Table~\ref{tab:silhouette} we present the silhouette values obtained using fine-tuned BERT.  As can be seen in the table, the best Spearman correlation coefficient does not correspond to the best Silhoutte score.  Moreover, the Silhouette scores are quite close to zero, which probably means that the clusters are not convex. 

The second hypothesis is connected with the difference in fine-tuning regimes employed in this research and the one conducted by \cite{giulianellilexical}. We use domain adaptation fine-tuning, proving its efficiency for a certain number of epochs, for both k-means (except for a small number of clusters) and affinity propagation. However, \cite{giulianellilexical} tried both diachronic fine-tuning (using the incremental fine-tuning technique first proposed by \cite{kim2014temporal}) and domain-specific fine-tuning, but concluded that none led to an improvement in the results. As it was already speculated in \cite{giulianellilexical}, using both training regimes at the same time might lead to too extensive fine-tuning and therefore over-fitting. Further, a more thorough study on influence of incremental fine-tuning on contextual embeddings models (such as BERT) should perhaps be conducted, since the effects might differ from the ones observed for static embeddings models. Finally, the domain-specific fine-tuning is conducted only for 1 to 3 epochs, which might be too few to improve the results on some corpora.

\begin{table}[!t]
\begin{tabular}{|l|r|r|}
\hline
\multicolumn{1}{|c|}{\textbf{Method}} & \textbf{Spearman} & \textbf{Silhouette} \\ 
\hline
\textbf{k-means, k=3}    & 0.392 & \textbf{0.105}\\
\textbf{k-means, k=5}    & 0.508 & 0.098\\
\textbf{k-means, k=7}    & 0.491 & 0.092\\
\textbf{k-means, k=10}   & 0.466 & 0.088\\
\textbf{k-means, k=100}  & {\em 0.337} & {\em 0.042} \\
\textbf{Affinity propagation} & \textbf{0.510} & 0.043 \\
\hline
\end{tabular}
\caption{Spearman correlation and Silhoutte score for clusterings obtained using fine-tuned BERT}
\label{tab:silhouette}
\vspace{-4mm}
\end{table}

\begin{figure}[t]
    \includegraphics[width=0.5\textwidth]{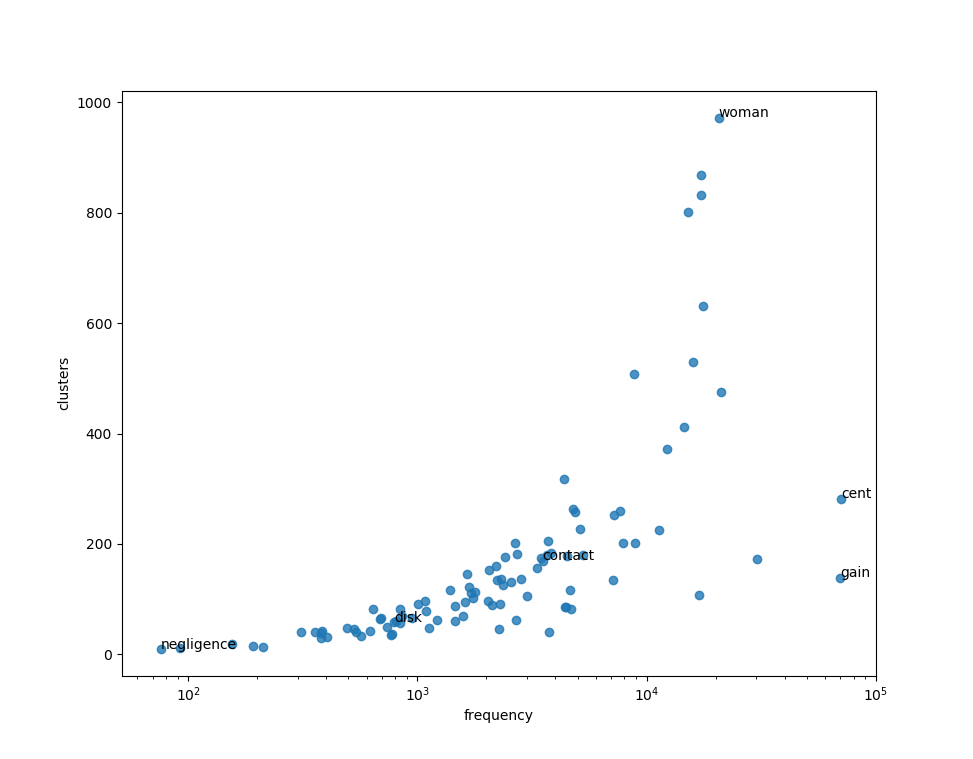}
    \caption{Number of clusters found by affinity propagation and frequency of a word in the 1960s and 1990s in COHA.}
    \label{fig:scatter_words}
    \vspace{-4mm}
\end{figure}

The difference in performance between k-means and affinity propagation could be partially explained by the different number of clusters in the two approaches. Affinity propagation, which performs the best, outputs a huge amount of clusters---160 on average.
The particular number of clusters found by affinity propagation for a word correlates strongly with the frequency of that word in the corpus with $r=0.875$, as is illustrated in Figure \ref{fig:scatter_words}. For instance, the word {\em woman} which occurred over 20k times in both time slices in COHA has the most number of clusters, 972, while {\em negligence}, occurring just 76 times has the least clusters, 10.

Thus, determining the optimal number of clusters for different words is not straightforward. We cannot claim that the clusters found by any of the methods we used can be interpreted as the different senses of a word or that they are even suitable for human interpretation.  Most probably, affinity propagation captures subtle differences in word usages rather than global semantic shift. Nevertheless, it works better than k-means with smaller and more intuitive number of clusters, since word sense induction and semantic shift detection are not the same task.

Affinity propagation usually produces rather skewed clustering, with a large number of small clusters containing only one or two data points, and thus can be used for outlier detection.  K-means is not suitable for this task since it uses a random initialisation and if an outlier is not initially selected as a potential centroid it may never be found.

To justify this claim we conducted an additional experiment and run k-means clustering on fine-tuned embeddings using k=100 or number of instances minus one for less frequent words.  This resulted in Pearson and Spearman rank correlations of 0.315 and 0.337, respectively, which is worse than {\em any} strategy presented in Table~\ref{tab:results} for fine-tuned embeddings, including averaging.  At the same time, the Silhouette score for this insufficient model is almost equal to the Silhoutte score for the best model, as shown in Table~\ref{tab:silhouette}.  Thus, Silhouette score fails to discriminate between the best and the worst model.

To conclude, clustering token embeddings using affinity propagation leads to the best results, with a Pearson correlation with human annotation of 0.56. To evaluate the success of our correlation results, we can use the value of the inter-rater agreement during the annotation process, which was 0.51, computed using the average of pair-wise Pearson correlations~\cite{Gulordava}. This highlights the difficulty of the task and the performance of the best method.

\section{Future Work}


One major issue in quantifying semantic change over time is the lack of gold standard data. In this work we used the 100 human-annotated words from ~\cite{Gulordava}; but to extend this study to a wider range of words for other corpora and other time slices we can construct datasets that simulates synthetic change as done in \cite{shoemark2019room}. This involves constructing "pseudowords" that are concatenations of two real words, and then injecting it in place of the real words with controlled probability distributions over a series of time slices. 

We also plan to investigate how the clusters found by the methods in this work can be used to interpret the different usages of a word in a specific time slice. The initial experiments on this subject have already been conducted with the two-stage clustering, which removes the smallest clusters, containing one or two instances. Thus, it allows to focus on a smaller number of the most representative clusters, which might be more suitable for human interpretation even though it does not yield the best result.  The initial check demonstrated that most of these clusters are interpretable, though some particular meaning can be spread among several clusters.

Our analysis hints that clustering BERT token embeddings for a word does not necessarily lead to sense-specific clusters. This conclusion is on par with~\cite{Coenen2019VisualizingAM}. Indeed, even though BERT is more adapted than ELMO~\cite{peters2018deep} or Flair NLP~\cite{Akbik2018ContextualSE} for word sense disambiguation~\cite{BERTsense19}, its ability do detect distinct word meanings has limitations. Thus, it would be interesting to extract only the semantic parts of the BERT embeddings to direct our analysis more towards word meaning and not word usage in general.

Clustering methods can be improved by taking into account multi-word expressions and named entities.  E.g. a company called Vector Security International appears only in 1990s time slice and this distorts semantic change calculations for word {\em vector}.  Such examples should be discarded.

Another interesting line of research would be to investigate the peculiar behaviour of the Silhouette score and try to find a better way for unsupervised cluster evaluation.  Though finding an universal evaluation measure might be very hard and using task-specific evaluation---as in this paper---could be a more reasonable strategy. 

\begin{acks}

We are grateful to Andrey Kutuzov for valuable discussions during this paper preparation. We also thank Pr. Alexandre Allauzen from ESPCI - Université Paris Dauphine and Pr. Asanobu Kitamoto from National Institute of Informatics (Tokyo) for their advises.

This work has been partly supported by the European Union’s Horizon 2020 research and innovation programme under grants 770299 (NewsEye) and 825153 (EMBEDDIA). 

\end{acks}

\bibliographystyle{ACM-Reference-Format}
\bibliography{bibliography}


\begin{thebibliography}{36}


\ifx \showCODEN    \undefined \def \showCODEN     #1{\unskip}     \fi
\ifx \showDOI      \undefined \def \showDOI       #1{#1}\fi
\ifx \showISBNx    \undefined \def \showISBNx     #1{\unskip}     \fi
\ifx \showISBNxiii \undefined \def \showISBNxiii  #1{\unskip}     \fi
\ifx \showISSN     \undefined \def \showISSN      #1{\unskip}     \fi
\ifx \showLCCN     \undefined \def \showLCCN      #1{\unskip}     \fi
\ifx \shownote     \undefined \def \shownote      #1{#1}          \fi
\ifx \showarticletitle \undefined \def \showarticletitle #1{#1}   \fi
\ifx \showURL      \undefined \def \showURL       {\relax}        \fi
\providecommand\bibfield[2]{#2}
\providecommand\bibinfo[2]{#2}
\providecommand\natexlab[1]{#1}
\providecommand\showeprint[2][]{arXiv:#2}

\bibitem[\protect\citeauthoryear{Aitchison}{Aitchison}{2001}]%
        {Aitchison}
\bibfield{author}{\bibinfo{person}{Jean Aitchison}.}
  \bibinfo{year}{2001}\natexlab{}.
\newblock \showarticletitle{Language Change: Progress Or Decay?}
\newblock In \bibinfo{booktitle}{\emph{Cambridge Approaches to Linguistics}}.
  \bibinfo{publisher}{Cambridge University Press},
  \bibinfo{address}{Cambridge}.
\newblock


\bibitem[\protect\citeauthoryear{Akbik, Blythe, and Vollgraf}{Akbik
  et~al\mbox{.}}{2018}]%
        {Akbik2018ContextualSE}
\bibfield{author}{\bibinfo{person}{Alan Akbik}, \bibinfo{person}{Duncan
  Blythe}, {and} \bibinfo{person}{Roland Vollgraf}.}
  \bibinfo{year}{2018}\natexlab{}.
\newblock \showarticletitle{Contextual string embeddings for sequence
  labeling}. In \bibinfo{booktitle}{\emph{Proceedings of the 27th International
  Conference on Computational Linguistics}}. \bibinfo{pages}{1638--1649}.
\newblock


\bibitem[\protect\citeauthoryear{Alagi{\'c}, {\v{S}}najder, and
  Pad{\'o}}{Alagi{\'c} et~al\mbox{.}}{2018}]%
        {alagic2018leveraging}
\bibfield{author}{\bibinfo{person}{Domagoj Alagi{\'c}}, \bibinfo{person}{Jan
  {\v{S}}najder}, {and} \bibinfo{person}{Sebastian Pad{\'o}}.}
  \bibinfo{year}{2018}\natexlab{}.
\newblock \showarticletitle{Leveraging lexical substitutes for unsupervised
  word sense induction}. In \bibinfo{booktitle}{\emph{Thirty-Second AAAI
  Conference on Artificial Intelligence}}.
\newblock


\bibitem[\protect\citeauthoryear{Amrami and Goldberg}{Amrami and
  Goldberg}{2019}]%
        {Amrami-WSI}
\bibfield{author}{\bibinfo{person}{Asaf Amrami} {and} \bibinfo{person}{Yoav
  Goldberg}.} \bibinfo{year}{2019}\natexlab{}.
\newblock \showarticletitle{Towards better substitution-based word sense
  induction}.
\newblock \bibinfo{journal}{\emph{arXiv preprint arXiv:1905.12598}}
  (\bibinfo{year}{2019}).
\newblock


\bibitem[\protect\citeauthoryear{Coenen, Reif, Yuan, Kim, Pearce, Vi{\'e}gas,
  and Wattenberg}{Coenen et~al\mbox{.}}{2019}]%
        {Coenen2019VisualizingAM}
\bibfield{author}{\bibinfo{person}{Andy Coenen}, \bibinfo{person}{Emily Reif},
  \bibinfo{person}{Ann Yuan}, \bibinfo{person}{Been Kim}, \bibinfo{person}{Adam
  Pearce}, \bibinfo{person}{Fernanda~B. Vi{\'e}gas}, {and}
  \bibinfo{person}{Martin Wattenberg}.} \bibinfo{year}{2019}\natexlab{}.
\newblock \showarticletitle{Visualizing and Measuring the Geometry of BERT}. In
  \bibinfo{booktitle}{\emph{NeurIPS}}.
\newblock


\bibitem[\protect\citeauthoryear{Devlin, Chang, Lee, and Toutanova}{Devlin
  et~al\mbox{.}}{2018}]%
        {devlin2018bert}
\bibfield{author}{\bibinfo{person}{Jacob Devlin}, \bibinfo{person}{Ming-Wei
  Chang}, \bibinfo{person}{Kenton Lee}, {and} \bibinfo{person}{Kristina
  Toutanova}.} \bibinfo{year}{2018}\natexlab{}.
\newblock \showarticletitle{Bert: Pre-training of deep bidirectional
  transformers for language understanding}.
\newblock \bibinfo{journal}{\emph{arXiv preprint arXiv:1810.04805}}
  (\bibinfo{year}{2018}).
\newblock


\bibitem[\protect\citeauthoryear{Dubossarsky, Hengchen, Tahmasebi, and
  Schlechtweg}{Dubossarsky et~al\mbox{.}}{2019}]%
        {dubossarsky2019time}
\bibfield{author}{\bibinfo{person}{Haim Dubossarsky}, \bibinfo{person}{Simon
  Hengchen}, \bibinfo{person}{Nina Tahmasebi}, {and} \bibinfo{person}{Dominik
  Schlechtweg}.} \bibinfo{year}{2019}\natexlab{}.
\newblock \showarticletitle{Time-Out: Temporal Referencing for Robust Modeling
  of Lexical Semantic Change}. In \bibinfo{booktitle}{\emph{Proceedings of the
  57th Annual Meeting of the Association for Computational Linguistics}}.
  \bibinfo{pages}{457--470}.
\newblock


\bibitem[\protect\citeauthoryear{Frermann and Lapata}{Frermann and
  Lapata}{2016}]%
        {frermann2016bayesian}
\bibfield{author}{\bibinfo{person}{Lea Frermann} {and} \bibinfo{person}{Mirella
  Lapata}.} \bibinfo{year}{2016}\natexlab{}.
\newblock \showarticletitle{A bayesian model of diachronic meaning change}.
\newblock \bibinfo{journal}{\emph{Transactions of the Association for
  Computational Linguistics}}  \bibinfo{volume}{4} (\bibinfo{year}{2016}),
  \bibinfo{pages}{31--45}.
\newblock


\bibitem[\protect\citeauthoryear{Frey and Dueck}{Frey and Dueck}{2007}]%
        {frey2007clustering}
\bibfield{author}{\bibinfo{person}{Brendan~J Frey} {and}
  \bibinfo{person}{Delbert Dueck}.} \bibinfo{year}{2007}\natexlab{}.
\newblock \showarticletitle{Clustering by passing messages between data
  points}.
\newblock \bibinfo{journal}{\emph{science}} \bibinfo{volume}{315},
  \bibinfo{number}{5814} (\bibinfo{year}{2007}), \bibinfo{pages}{972--976}.
\newblock


\bibitem[\protect\citeauthoryear{Gillani and Levy}{Gillani and Levy}{2019}]%
        {GillaniDynamicPerception19}
\bibfield{author}{\bibinfo{person}{Nabeel Gillani} {and} \bibinfo{person}{Roger
  Levy}.} \bibinfo{year}{2019}\natexlab{}.
\newblock \showarticletitle{Simple dynamic word embeddings for mapping
  perceptions in the public sphere}. \bibinfo{pages}{94--99}.
\newblock
\urldef\tempurl%
\url{https://doi.org/10.18653/v1/W19-2111}
\showDOI{\tempurl}


\bibitem[\protect\citeauthoryear{Giulianelli}{Giulianelli}{2019}]%
        {giulianellilexical}
\bibfield{author}{\bibinfo{person}{Mario Giulianelli}.}
  \bibinfo{year}{2019}\natexlab{}.
\newblock \bibinfo{booktitle}{\emph{Lexical Semantic Change Analysis with
  Contextualised Word Representations}}.
\newblock \bibinfo{publisher}{University of Amsterdam - Institute for logic,
  Language and computation}.
\newblock


\bibitem[\protect\citeauthoryear{Goldberg and Orwant}{Goldberg and
  Orwant}{2013}]%
        {goldberg2013dataset}
\bibfield{author}{\bibinfo{person}{Yoav Goldberg} {and} \bibinfo{person}{Jon
  Orwant}.} \bibinfo{year}{2013}\natexlab{}.
\newblock \showarticletitle{A dataset of syntactic-ngrams over time from a very
  large corpus of english books}. In \bibinfo{booktitle}{\emph{Second Joint
  Conference on Lexical and Computational Semantics}}.
  \bibinfo{pages}{241--247}.
\newblock


\bibitem[\protect\citeauthoryear{Gulordava and Baroni}{Gulordava and
  Baroni}{2011}]%
        {Gulordava}
\bibfield{author}{\bibinfo{person}{Kristina Gulordava} {and}
  \bibinfo{person}{Marco Baroni}.} \bibinfo{year}{2011}\natexlab{}.
\newblock \showarticletitle{A distributional similarity approach to the
  detection of semantic change in the Google Books Ngram corpus.}
  \bibinfo{publisher}{Association for Computational Linguistics},
  \bibinfo{pages}{67--71}.
\newblock
\urldef\tempurl%
\url{http://aclweb.org/anthology/W11-2508}
\showURL{%
\tempurl}


\bibitem[\protect\citeauthoryear{Hilpert and Gries}{Hilpert and Gries}{2008}]%
        {hilpert2008assessing}
\bibfield{author}{\bibinfo{person}{Martin Hilpert} {and}
  \bibinfo{person}{Stefan~Th Gries}.} \bibinfo{year}{2008}\natexlab{}.
\newblock \showarticletitle{Assessing frequency changes in multistage
  diachronic corpora: Applications for historical corpus linguistics and the
  study of language acquisition}.
\newblock \bibinfo{journal}{\emph{Literary and Linguistic Computing}}
  \bibinfo{volume}{24}, \bibinfo{number}{4} (\bibinfo{year}{2008}),
  \bibinfo{pages}{385--401}.
\newblock


\bibitem[\protect\citeauthoryear{Howard and Ruder}{Howard and Ruder}{2018}]%
        {howard2018universal}
\bibfield{author}{\bibinfo{person}{Jeremy Howard} {and}
  \bibinfo{person}{Sebastian Ruder}.} \bibinfo{year}{2018}\natexlab{}.
\newblock \showarticletitle{Universal language model fine-tuning for text
  classification}.
\newblock \bibinfo{journal}{\emph{arXiv preprint arXiv:1801.06146}}
  (\bibinfo{year}{2018}).
\newblock


\bibitem[\protect\citeauthoryear{Hu, Li, and Liang}{Hu et~al\mbox{.}}{2019}]%
        {hu-etal-2019-diachronic}
\bibfield{author}{\bibinfo{person}{Renfen Hu}, \bibinfo{person}{Shen Li}, {and}
  \bibinfo{person}{Shichen Liang}.} \bibinfo{year}{2019}\natexlab{}.
\newblock \showarticletitle{Diachronic Sense Modeling with Deep Contextualized
  Word Embeddings: An Ecological View}. In
  \bibinfo{booktitle}{\emph{Proceedings of the 57th Annual Meeting of the
  Association for Computational Linguistics}}. \bibinfo{publisher}{Association
  for Computational Linguistics}, \bibinfo{address}{Florence, Italy},
  \bibinfo{pages}{3899--3908}.
\newblock
\urldef\tempurl%
\url{https://doi.org/10.18653/v1/P19-1379}
\showDOI{\tempurl}


\bibitem[\protect\citeauthoryear{Juola}{Juola}{2003}]%
        {juola2003time}
\bibfield{author}{\bibinfo{person}{Patrick Juola}.}
  \bibinfo{year}{2003}\natexlab{}.
\newblock \showarticletitle{The time course of language change}.
\newblock \bibinfo{journal}{\emph{Computers and the Humanities}}
  \bibinfo{volume}{37}, \bibinfo{number}{1} (\bibinfo{year}{2003}),
  \bibinfo{pages}{77--96}.
\newblock


\bibitem[\protect\citeauthoryear{Kaji and Kobayashi}{Kaji and
  Kobayashi}{2017}]%
        {kaji2017incremental}
\bibfield{author}{\bibinfo{person}{Nobuhiro Kaji} {and} \bibinfo{person}{Hayato
  Kobayashi}.} \bibinfo{year}{2017}\natexlab{}.
\newblock \showarticletitle{Incremental skip-gram model with negative
  sampling}.
\newblock \bibinfo{journal}{\emph{arXiv preprint arXiv:1704.03956}}
  (\bibinfo{year}{2017}).
\newblock


\bibitem[\protect\citeauthoryear{Kim, Chiu, Hanaki, Hegde, and Petrov}{Kim
  et~al\mbox{.}}{2014}]%
        {kim2014temporal}
\bibfield{author}{\bibinfo{person}{Yoon Kim}, \bibinfo{person}{Yi-I Chiu},
  \bibinfo{person}{Kentaro Hanaki}, \bibinfo{person}{Darshan Hegde}, {and}
  \bibinfo{person}{Slav Petrov}.} \bibinfo{year}{2014}\natexlab{}.
\newblock \showarticletitle{Temporal analysis of language through neural
  language models}.
\newblock \bibinfo{journal}{\emph{arXiv preprint arXiv:1405.3515}}
  (\bibinfo{year}{2014}).
\newblock


\bibitem[\protect\citeauthoryear{Kulkarni, Al-Rfou, Perozzi, and
  Skiena}{Kulkarni et~al\mbox{.}}{2015}]%
        {kulkarni2015statistically}
\bibfield{author}{\bibinfo{person}{Vivek Kulkarni}, \bibinfo{person}{Rami
  Al-Rfou}, \bibinfo{person}{Bryan Perozzi}, {and} \bibinfo{person}{Steven
  Skiena}.} \bibinfo{year}{2015}\natexlab{}.
\newblock \showarticletitle{Statistically significant detection of linguistic
  change}. In \bibinfo{booktitle}{\emph{Proceedings of the 24th International
  Conference on World Wide Web}}. International World Wide Web Conferences
  Steering Committee, \bibinfo{pages}{625--635}.
\newblock


\bibitem[\protect\citeauthoryear{Kutuzov}{Kutuzov}{2020}]%
        {kutuzov2020diachronic}
\bibfield{author}{\bibinfo{person}{Andrey Kutuzov}.}
  \bibinfo{year}{2020}\natexlab{}.
\newblock \showarticletitle{Diachronic contextualized embeddings and semantic
  shifts}.
\newblock


\bibitem[\protect\citeauthoryear{Kutuzov, Kuzmenko, and Pivovarova}{Kutuzov
  et~al\mbox{.}}{2017}]%
        {kutuzov2017clustering}
\bibfield{author}{\bibinfo{person}{Andrey Kutuzov}, \bibinfo{person}{Elizaveta
  Kuzmenko}, {and} \bibinfo{person}{Lidia Pivovarova}.}
  \bibinfo{year}{2017}\natexlab{}.
\newblock \showarticletitle{Clustering of Russian Adjective-Noun Constructions
  Using Word Embeddings}. In \bibinfo{booktitle}{\emph{Proceedings of the 6th
  Workshop on Balto-Slavic Natural Language Processing}}.
  \bibinfo{pages}{3--13}.
\newblock


\bibitem[\protect\citeauthoryear{Kutuzov, {\O}vrelid, Szymanski, and
  Velldal}{Kutuzov et~al\mbox{.}}{2018}]%
        {kutuzov2018diachronic}
\bibfield{author}{\bibinfo{person}{Andrey Kutuzov}, \bibinfo{person}{Lilja
  {\O}vrelid}, \bibinfo{person}{Terrence Szymanski}, {and}
  \bibinfo{person}{Erik Velldal}.} \bibinfo{year}{2018}\natexlab{}.
\newblock \showarticletitle{Diachronic word embeddings and semantic shifts: a
  survey}.
\newblock \bibinfo{journal}{\emph{arXiv preprint arXiv:1806.03537}}
  (\bibinfo{year}{2018}).
\newblock


\bibitem[\protect\citeauthoryear{Martinc, Novak, and Pollak}{Martinc
  et~al\mbox{.}}{2019}]%
        {martinc2019leveraging}
\bibfield{author}{\bibinfo{person}{Matej Martinc}, \bibinfo{person}{Petra~Kralj
  Novak}, {and} \bibinfo{person}{Senja Pollak}.}
  \bibinfo{year}{2019}\natexlab{}.
\newblock \showarticletitle{Leveraging Contextual Embeddings for Detecting
  Diachronic Semantic Shift}.
\newblock \bibinfo{journal}{\emph{arXiv preprint arXiv:1912.01072}}
  (\bibinfo{year}{2019}).
\newblock


\bibitem[\protect\citeauthoryear{Mikolov, Chen, Corrado, and Dean}{Mikolov
  et~al\mbox{.}}{2013}]%
        {mikolov2013efficient}
\bibfield{author}{\bibinfo{person}{Tomas Mikolov}, \bibinfo{person}{Kai Chen},
  \bibinfo{person}{Greg Corrado}, {and} \bibinfo{person}{Jeffrey Dean}.}
  \bibinfo{year}{2013}\natexlab{}.
\newblock \showarticletitle{Efficient estimation of word representations in
  vector space}.
\newblock \bibinfo{journal}{\emph{arXiv preprint arXiv:1301.3781}}
  (\bibinfo{year}{2013}).
\newblock


\bibitem[\protect\citeauthoryear{Peng, Li, Song, and Liu}{Peng
  et~al\mbox{.}}{2017}]%
        {peng2017incrementally}
\bibfield{author}{\bibinfo{person}{Hao Peng}, \bibinfo{person}{Jianxin Li},
  \bibinfo{person}{Yangqiu Song}, {and} \bibinfo{person}{Yaopeng Liu}.}
  \bibinfo{year}{2017}\natexlab{}.
\newblock \showarticletitle{Incrementally learning the hierarchical softmax
  function for neural language models}. In
  \bibinfo{booktitle}{\emph{Thirty-First AAAI Conference on Artificial
  Intelligence}}.
\newblock


\bibitem[\protect\citeauthoryear{Pennington, Socher, and Manning}{Pennington
  et~al\mbox{.}}{2014}]%
        {Pennington14GLOVE}
\bibfield{author}{\bibinfo{person}{Jeffrey Pennington},
  \bibinfo{person}{Richard Socher}, {and} \bibinfo{person}{Christopher~D.
  Manning}.} \bibinfo{year}{2014}\natexlab{}.
\newblock \showarticletitle{GloVe: Global Vectors for Word Representation}. In
  \bibinfo{booktitle}{\emph{Empirical Methods in Natural Language Processing
  (EMNLP)}}. \bibinfo{pages}{1532--1543}.
\newblock


\bibitem[\protect\citeauthoryear{Peters, Neumann, Iyyer, Gardner, Clark, Lee,
  and Zettlemoyer}{Peters et~al\mbox{.}}{2018}]%
        {peters2018deep}
\bibfield{author}{\bibinfo{person}{Matthew~E Peters}, \bibinfo{person}{Mark
  Neumann}, \bibinfo{person}{Mohit Iyyer}, \bibinfo{person}{Matt Gardner},
  \bibinfo{person}{Christopher Clark}, \bibinfo{person}{Kenton Lee}, {and}
  \bibinfo{person}{Luke Zettlemoyer}.} \bibinfo{year}{2018}\natexlab{}.
\newblock \showarticletitle{Deep contextualized word representations}.
\newblock \bibinfo{journal}{\emph{arXiv preprint arXiv:1802.05365}}
  (\bibinfo{year}{2018}).
\newblock


\bibitem[\protect\citeauthoryear{Rosenfeld and Erk}{Rosenfeld and Erk}{2018}]%
        {rosenfeld2018deep}
\bibfield{author}{\bibinfo{person}{Alex Rosenfeld} {and}
  \bibinfo{person}{Katrin Erk}.} \bibinfo{year}{2018}\natexlab{}.
\newblock \showarticletitle{Deep neural models of semantic shift}. In
  \bibinfo{booktitle}{\emph{Proceedings of the 2018 Conference of the North
  American Chapter of the Association for Computational Linguistics: Human
  Language Technologies, Volume 1 (Long Papers)}}. \bibinfo{pages}{474--484}.
\newblock


\bibitem[\protect\citeauthoryear{Shoemark, Liza, Nguyen, Hale, and
  McGillivray}{Shoemark et~al\mbox{.}}{2019}]%
        {shoemark2019room}
\bibfield{author}{\bibinfo{person}{Philippa Shoemark},
  \bibinfo{person}{Farhana~Ferdousi Liza}, \bibinfo{person}{Dong Nguyen},
  \bibinfo{person}{Scott Hale}, {and} \bibinfo{person}{Barbara McGillivray}.}
  \bibinfo{year}{2019}\natexlab{}.
\newblock \showarticletitle{Room to Glo: A Systematic Comparison of Semantic
  Change Detection Approaches with Word Embeddings}. In
  \bibinfo{booktitle}{\emph{Proceedings of the 2019 Conference on Empirical
  Methods in Natural Language Processing and the 9th International Joint
  Conference on Natural Language Processing (EMNLP-IJCNLP)}}.
  \bibinfo{pages}{66--76}.
\newblock


\bibitem[\protect\citeauthoryear{Tahmasebi, Borin, and Jatowt}{Tahmasebi
  et~al\mbox{.}}{2018}]%
        {tahmasebi2018survey}
\bibfield{author}{\bibinfo{person}{Nina Tahmasebi}, \bibinfo{person}{Lars
  Borin}, {and} \bibinfo{person}{Adam Jatowt}.}
  \bibinfo{year}{2018}\natexlab{}.
\newblock \showarticletitle{Survey of Computational Approaches to Diachronic
  Conceptual Change}.
\newblock \bibinfo{journal}{\emph{arXiv preprint arXiv:1811.06278}}
  (\bibinfo{year}{2018}).
\newblock


\bibitem[\protect\citeauthoryear{Tang}{Tang}{2018}]%
        {tang_2018}
\bibfield{author}{\bibinfo{person}{Xuri Tang}.}
  \bibinfo{year}{2018}\natexlab{}.
\newblock \showarticletitle{A state-of-the-art of semantic change computation}.
\newblock \bibinfo{journal}{\emph{Natural Language Engineering}}
  \bibinfo{volume}{24}, \bibinfo{number}{5} (\bibinfo{year}{2018}),
  \bibinfo{pages}{649–676}.
\newblock
\urldef\tempurl%
\url{https://doi.org/10.1017/S1351324918000220}
\showDOI{\tempurl}


\bibitem[\protect\citeauthoryear{Vaswani, Shazeer, Parmar, Uszkoreit, Jones,
  Gomez, Kaiser, and Polosukhin}{Vaswani et~al\mbox{.}}{2017}]%
        {vaswani2017attention}
\bibfield{author}{\bibinfo{person}{Ashish Vaswani}, \bibinfo{person}{Noam
  Shazeer}, \bibinfo{person}{Niki Parmar}, \bibinfo{person}{Jakob Uszkoreit},
  \bibinfo{person}{Llion Jones}, \bibinfo{person}{Aidan~N Gomez},
  \bibinfo{person}{{\L}ukasz Kaiser}, {and} \bibinfo{person}{Illia
  Polosukhin}.} \bibinfo{year}{2017}\natexlab{}.
\newblock \showarticletitle{Attention is all you need}. In
  \bibinfo{booktitle}{\emph{Advances in Neural Information Processing
  Systems}}. \bibinfo{pages}{5998--6008}.
\newblock


\bibitem[\protect\citeauthoryear{Wiedemann, Remus, Chawla, and
  Biemann}{Wiedemann et~al\mbox{.}}{2019}]%
        {BERTsense19}
\bibfield{author}{\bibinfo{person}{Gregor Wiedemann}, \bibinfo{person}{Steffen
  Remus}, \bibinfo{person}{Avi Chawla}, {and} \bibinfo{person}{Chris Biemann}.}
  \bibinfo{year}{2019}\natexlab{}.
\newblock \showarticletitle{Does BERT Make Any Sense? Interpretable Word Sense
  Disambiguation with Contextualized Embeddings}.
\newblock


\bibitem[\protect\citeauthoryear{Wu, Schuster, Chen, Le, Norouzi, Macherey,
  Krikun, Cao, Gao, Macherey, et~al\mbox{.}}{Wu et~al\mbox{.}}{2016}]%
        {wu2016google}
\bibfield{author}{\bibinfo{person}{Yonghui Wu}, \bibinfo{person}{Mike
  Schuster}, \bibinfo{person}{Zhifeng Chen}, \bibinfo{person}{Quoc~V Le},
  \bibinfo{person}{Mohammad Norouzi}, \bibinfo{person}{Wolfgang Macherey},
  \bibinfo{person}{Maxim Krikun}, \bibinfo{person}{Yuan Cao},
  \bibinfo{person}{Qin Gao}, \bibinfo{person}{Klaus Macherey}, {et~al\mbox{.}}}
  \bibinfo{year}{2016}\natexlab{}.
\newblock \showarticletitle{Google's neural machine translation system:
  Bridging the gap between human and machine translation}.
\newblock \bibinfo{journal}{\emph{arXiv preprint arXiv:1609.08144}}
  (\bibinfo{year}{2016}).
\newblock


\bibitem[\protect\citeauthoryear{Zhang, Jatowt, Bhowmick, and Tanaka}{Zhang
  et~al\mbox{.}}{2016}]%
        {zhang2016past}
\bibfield{author}{\bibinfo{person}{Yating Zhang}, \bibinfo{person}{Adam
  Jatowt}, \bibinfo{person}{Sourav~S Bhowmick}, {and} \bibinfo{person}{Katsumi
  Tanaka}.} \bibinfo{year}{2016}\natexlab{}.
\newblock \showarticletitle{The past is not a foreign country: Detecting
  semantically similar terms across time}.
\newblock \bibinfo{journal}{\emph{IEEE Transactions on Knowledge and Data
  Engineering}} \bibinfo{volume}{28}, \bibinfo{number}{10}
  (\bibinfo{year}{2016}), \bibinfo{pages}{2793--2807}.
\newblock


\end{thebibliography}


\end{document}